\documentclass[runningheads]{llncs}
\usepackage[T1]{fontenc}
\usepackage{graphicx}
\usepackage{float}
\usepackage[linesnumbered,ruled]{algorithm2e}
\usepackage{algpseudocode}
\usepackage{amsmath}
\usepackage{cite}
\usepackage{graphicx}
\usepackage{footnote}
\usepackage{bbding}
\usepackage{pifont}
\usepackage{wasysym}
\usepackage{amssymb}
\usepackage{multirow}
\usepackage[table,xcdraw]{xcolor}
\usepackage{diagbox}
\usepackage[flushleft]{threeparttable}
\makesavenoteenv{tabular}
\usepackage{tabu}
\usepackage{longtable}
\usepackage{cite}
\usepackage{epstopdf}
\usepackage{color}
\usepackage{enumitem}
\usepackage{booktabs}
\usepackage{subfigure}
\setlist{parsep=0pt,listparindent=\parindent}
\usepackage{caption}
\newtheorem{myDef}{Definition}

\usepackage{ulem}

\usepackage[cmintegrals]{newtxmath}

\usepackage[misc]{ifsym}

\begin{document}
\title{Discover Important Paths in the Knowledge Graph Based on Dynamic Relation Confidence}
\titlerunning{Discover important paths in the KG based on DC-path}
\author{Shanqing Yu\inst{1,2} \textsuperscript{(\Letter)} \and
        Yijun Wu\inst{1,2} \and
        Ran Gan\inst{1,2} \and  \\
        Jiajun Zhou\inst{1,2} \and
        Ziwan Zheng\inst{3} \and 
        Qi Xuan\inst{1,2}}
\authorrunning{S.Yu et al.}
\institute{
    Institute of Cyberspace Security, Zhejiang University of Technology, \\Hangzhou 310023, China \\ \and
    College of Information Engineering, Zhejiang University of Technology, \\Hangzhou 310023, China\\ \and Zhejiang Police College,Hangzhou 310053,China\\
    \email{yushanqing@zjut.edu.cn}}
\maketitle              % typeset the header of the contribution
\begin{abstract}
Most of the existing knowledge graphs are not usually complete and can be complemented by some reasoning algorithms. The reasoning method based on path features is widely used in the field of knowledge graph reasoning and completion on account of that its have strong interpretability. However, reasoning methods based on path features still have several problems in the following aspects: Path search isinefficient, insufficient paths for sparse tasks and some paths are not helpful for reasoning tasks. In order to solve the above problems, this paper proposes a method called DC-Path that combines dynamic relation confidence and other indicators to evaluate path features, and then guide path search, finally conduct relation reasoning. Experimental result show that compared with the existing relation reasoning algorithm, this method can select the most representative features in the current reasoning task from the knowledge graph and achieve better performance on the current relation reasoning task.

\end{abstract}
\section{Introduction}
%\subsection{A Subsection Sample}
Knowledge graph(KG) can be considered as a variant of semantic network with added constraints, or a programmatic way to model a knowledge domain. Knowledge graph reasoning, which focuses on inferring new unknown knowledge from the existing KG, has been widely deployed in KG completion. For knowledge reasoning, commonly used methods concentrate on representation learning, rule, graph structure, and deep learning methods.

The KG reasoning method based on path features is an important part of the graph structure reasoning methods. This kind of method usually includes path search and reasoning. Since the path features is composed of the relation sequences in the KG, it has strong interpretability. The KG reasoning method based on path features can be traced back to the path ranking algorithm(PRA) proposed by Lao et al.~\cite{lao2010fast}, which extracts the relation sequences between entities as features. Later, a series of improved algorithms for path search has emerged.

This paper observes and analyzes a series of typical path reasoning algorithms. In general, there are three problems that cannot be avoided in the path search and reasoning: 1. There are too many relations in a large KG, resulting in an inefficient path search. 2. Due to the sparsity of the KG, some reasoning tasks cannot find enough path features for reasoning. 3.Some path features are not relevant to the current reasoning task, so they are not helpful for reasoning.

Although some algorithms have noticed these problems and made effective improvements, there are still some shortcomings. Lao et al.~\cite{lao2011random} took data-based path walks to improve the efficiency of path search, but it only evaluates and filters the path features at the end of the path search. Gardner M et al. proposed the SFE algorithm~\cite{gardner2015efficient} which divides the path search process into two subgraphs and searches for the intermediate entity at the same time, thereby improving the efficiency of path search. In addition, it binarizes the probability matrix to reduce the calculation. However, it still cannot choose the path related to the reasoning task and only improves efficiency. Xiong W et al. proposed the DeepPath algorithm~\cite{xiong2017deeppath}, which applies reinforcement learning to search paths. Its disadvantage is that the method of reinforcement learning depends on the quality of the embedding method used. Meanwhile, the reinforcement learning network needs to be pre-trained which consumes more time.

Based on the above, this paper proposes a method that uses dynamic relation and path confidence to evaluate the path and guide the path search. Its characteristic lies in dynamically evaluating relations and paths during the path search. In the whole search process, with the search strategy is continuously optimized, the search space is continuously reduced to the area most relevant to the reasoning task. Finally, the path is selected according to the path confidence and other indicators to retain the most important path features for the reasoning task. The main contributions of this paper are as follows:
 \begin{itemize}
\item [$\bullet$] We define the dynamic evaluation indicators to evaluate the quality of relation and path in KG.
This method includes path search, path selection, and finally perform relation reasoning tasks based on dynamic confidence indicators is called Dynamic confidence path(DC-Path).
\item [$\bullet$] We define dynamic relation confidence to guide the path search and narrow the path search space. Experiments show that path search through DC-Path can more effectively find the most important path for the current reasoning task.
\item [$\bullet$] We use different strategies for path selection to observe its impact on the reasoning results and discover which paths play a decisive role in the reasoning task.
\end{itemize}

The rest of this paper is composed as follows: Section2 briefly introduces related work about KG reasoning algorithms. Section3 introduces our method, and Section4 show our experiments and analysis. Section5 summarizes the full text.

\section{Related Work}
\subsection{Reasoning Method Based On Path Features}

Path ranking algorithm (PRA) is the earliest and classic algorithm based on path features reasoning in KG. Lao et al. further improved the PRA algorithm in paper~\cite{lao2015learning}. In this paper, a method of adding a reverse random walk is proposed to expand the original walking strategy, and a path containing constant is added to the path features. The DeepPath algorithm applies reinforcement learning to path search for the first time, and its core ideas are as follows: Get the current state according to the embedding of entities, and select which relation to search according to an action matrix. Set three types of rewards to continuously strengthen the strategy of walking, and finally extract the path features with the best performance. It improves the accuracy of relation reasoning tasks and uses fewer path features than PRA.After that, some new reinforcement learning methods for graph reasoning or path search were proposed~\cite{das2017go,fu2019collaborative,lin2018multi,shen2018m}. 

\subsection{The Reasoning Method Based On Representation Learning}
The reasoning method based on representation learning maps the entities and relations in the KG to a low-dimensional space and set a score function to evaluate the correctness of a triple. Translation models are typical KG reasoning algorithms based on representation learning such as TransE~\cite{bordes2013translating} and its series of improved TransH~\cite{wang2014knowledge}, TransR~\cite{TransR}, and TransD~\cite{ji2015knowledge}.
Another type of KG reasoning method that represents learning is the semantic matching model, and its typical algorithms include Analogy~\cite{liu2017analogical}. 
The advantage of this type of method is that after completing the embedding of entities and relations, reasoning can be performed efficiently through the scoring function.
Compared with the reasoning method based on path features, the reasoning performance is better when facing sparse KG.In recent years, more KG reasoning methods based on representation learning have been proposed~\cite{balavzevic2019tucker,sun2019rotate,xue2018expanding,zhang2019quaternion,zhang2019interaction}.

\subsection{Reasoning Method Based On Association Rules}

Association rule mining is another type of KG reasoning method. Association rules were first proposed for shopping analysis to indicate the shopping association in the market. There are also many association rules in KG. AMIE is a typical knowledge graph association rule mining system~\cite{galarraga2013amie}.In the AMIE system, rule confidence based partial completeness assumption is proposed to replace the traditional indicators in the field of original rule mining. It does not assume any fact that does not appear in the KG but assumes that it is missing. This inspired us to make similar confidence definitions for the paths and relations in the KG. At the same time, AMIE+~\cite{galarraga2015rule} also proposed a prediction method based on the partial completeness assumption confidence of association rules, which has achieved a remarkable effect on the yago3 dataset. Many rule-based KG reasoning methods have been proposed in recent years~\cite{guo2017knowledge,yang2017differentiable}.

\subsection{Reasoning Method Based On Neural Network}
Recent studies have shown that the neural network coding model has a good effect on the completion of the KG~\cite{ji2020survey}. Encoding models with linear/bilinear blocks can also be modeled by neural networks, such as NAM~\cite{liu2016probabilistic}. CNN are utilized for learning deep expressive features in recent years, its representative algorithms are ConvE~\cite{dettmers2017convolutional}, ConvKB ~\cite{nguyen2017novel}and HypER~\cite{balavzevic2019hypernetwork}.GNN encoder~\cite{jung2021learning}is also used to reasoning. 

\section{Method}

In this section, we propose a path search algorithm based on dynamic relation confidence, to search effective path features which are further used for relation reasoning.In general, the dynamic confidence of the relation will determine the search strategy.

\begin{figure}
\centerline{\includegraphics[width=0.98\linewidth]{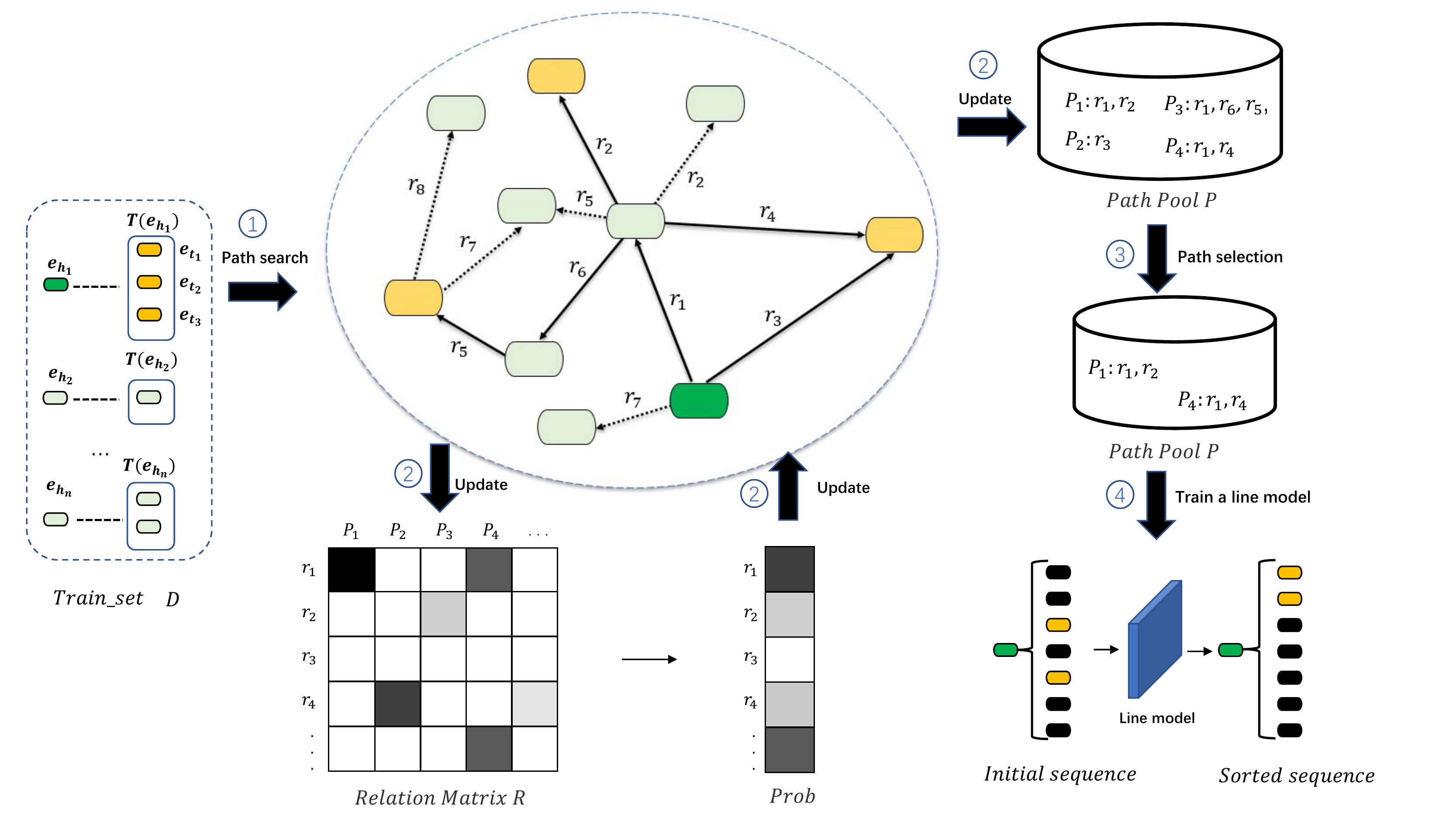}}
\caption{The framework of path search and reasoning. The workflow proceeds as follows: 1) traverse the positive sample entity pairs in turn for path search; 2) update the path pool and the relation confidence matrix, and adjust the path search strategies; 3) make path selection based on dynamic confidence and pairs coverage; 4) train linear regression model using the final path features to perform relation reasoning tasks.} \label{fig:frame}
\end{figure}

For a specific reasoning task, if a relation appears frequently in high-quality path features, we will gradually increase the search probability of it and vice versa. In doing so, we can narrow the search space to be more relevant to the target task, and obtain the most representative path features.The framework of path search and reasoning is shown in Fig~\ref{fig:frame}.Firstly, traverse the entity pairs in the training set for path search, during which both the path pool $\mathbf{P}_{l}$ and the relation matrix $C$ are dynamically updated to adjust the search probability. The path pool saves all the currently searched path features and their dynamic path confidence and the relation matrix saves the path dynamic confidence that each relation has participated in.After obtaining all the path features via traverse, we conduct path selection based on dynamic confidence and pairs coverage. Finally, we train a simple linear regression model using the final path features to perform relation reasoning tasks.

\subsection{KG Preprocessing}

Considering the sparsity and incompleteness of the existing KG, we conduct preliminary processing via inverse relation generation which is commonly used to expand the KG.
Specifically, we generate the inverse relation triples for the existing triples and add them to the KG, which can alleviate the problem that some reasoning tasks cannot find enough paths for reasoning due to the sparsity of KG.  

For example, we will add an inverse relation triple $\left(t,hasspouse^{-1},h\right)$ for the triple $\left(h,hasspouse,t\right)$. 
However, instead of adding an inverse relation triple for each triple directly, we first evaluate the relation in the KG. If a relation is frequently connected to the same tail entity through different head entities, we do not add an inverse relation triple for it.First of all, the triples composed of such a public entity usually represent common sense in the KG. In addition, adding such triples will generate an entity with a larger out-degree, which will affect the efficiency of path search.
Meanwhile, we do not remove redundant relations from semantic information, the reason is that the relation in the KG is extremely incomplete.

\subsection{Path Evaluation}\label{sec:path-eva}
Although the addition of inverse relation triples helps us expand the path features, it makes the existing path features more miscellaneous.
To evaluate the quality of various path features and select effective ones for relation reasoning, we define several evaluation measures as below: path-entity support, path count, path confidence, and entity pair coverage.
Notably, different from association rules, we only focus on the path and use dynamic confidence for approximate representation. 
\begin{myDef}
  (\textbf{Path-entity support})  For a given head entity $e_h$, tail entity $e_t$ and path feature $P=r_1,r_2,...,r_l$, the path-entity support is the number of instances of target entity pairs $(e_h, e_t)$ satisfying the path constraint $P$ in the KG:
  % \begin{equation}
  %   {support}\left(e_{h}, e_{t},{p}\right)=\exists e_{h} \stackrel{r_{1}}{\rightarrow} e_{i}  \stackrel{r_{2}}{\rightarrow} \cdots \cdot \stackrel{r_{l}}{\rightarrow} e_{t},
  % \end{equation}
  \begin{equation}
    \begin{array}{c}
      \textsf{support}\left(e_{h}, e_{t},{P}\right) = \text{the number of } \\
      \{e_{h} \stackrel{r_{1}}{\rightarrow} e_{i}  \stackrel{r_{2}}{\rightarrow} \cdots \cdot \stackrel{r_{l}}{\rightarrow} e_{t}\},
    \end{array}
  \end{equation}
where $e_{i}$ is an arbitrary entity in the KG.
\end{myDef}
Under the constraint of a specific path feature, the head entity can reach the tail entity through different entity sequences, so this indicator is usually greater than one.

\begin{myDef}
  (\textbf{Path count}) For a given head entity $e_h$ and path feature $P=r_1,r_2,...,r_l$, the path count is the number of instances of any tail entity that a head entity can reach under a specific path constraint $P$ in the KG:
  \begin{equation}
    \begin{array}{c}
      \textsf{count} \left(e_h, P\right) = \text{the number of } \\
      \{e_{h} \stackrel{r_{1}}{\rightarrow} e_{i}  \stackrel{r_{2}}{\rightarrow} \cdots \cdot \stackrel{r_{l}}{\rightarrow} e_{j}\},
    \end{array}
  \end{equation}
  where $e_{j}$ is an arbitrary entity in the KG.
\end{myDef}

\begin{myDef}
  (\textbf{Entity pair cover}) For a given head entity $e_h$, tail entity $e_t$ and a path feature $P=r_1,r_2,...,r_l$, the entity pair cover indicates whether the current entity pair $(e_h, e_t)$ meets the path constraint $P$:
  \begin{equation}
    \textsf{cover}\left(e_{h}, e_{t}, {P}\right)=\left\{
      \begin{aligned}
        1, \  & \textsf{support}\left(e_{h}, e_{t},{P}\right)>=1 \\
        0, \  & \textsf{support}\left(e_{h}, e_{t},{P}\right)=0  \\
      \end{aligned}
    \right .
  \end{equation} 
\end{myDef}

\begin{myDef}
  (\textbf{Path confidence}) For a given head entity $e_h$, tail entity $e_t$ and path feature $P=r_1,r_2,...,r_l$, the path confidence indicates the proportion of the target entity pairs $(e_h, e_t)$ to all entity pairs starting from the target head entity $e_h$ under a specific path constraint $P$:
  \begin{equation}
    \textsf{confidence}\left(P\right)=\frac{\sum_{i=1}^{|D|} \textsf{support}\left(e_{h_{i}}, e_{t_{i}}, P\right)}{\sum_{i=1}^{|D|} \textsf{count}\left(e_{h_{i}}, P\right)}, 
  \end{equation}
  where $|D|$ represents the total number of entity pairs in the training set.
\end{myDef}
The path confidence measures the overall reliability of a path in all entity pairs in the training set. When this value is 1, it means that the head entity of the positive sample can walk to the correct tail entity under the path constraint.

\begin{myDef}
  (\textbf{Entity pair coverage}) For a given head entity $e_h$, tail entity $e_t$ and path feature $P=r_1,r_2,...,r_l$, entity pair coverage represents the proportion of all entity pairs $(e_h, e_t)$ in the training set that satisfy the specific path constraint $P$:
  \begin{equation}
    \textsf{coverage}\left(P\right)=\frac{\sum_{i=1}^{|D|} \textsf{cover}\left(e_{h_{i}}, e_{t_{i}}, {P}\right)}{|D|}
  \end{equation}
\end{myDef}

\subsection{Path Search and Strategy Update}
The indicators proposed in Sec.~\ref{sec:path-eva} can evaluate path feature well, but they cannot be calculated during path search, and only can be calculated after path search, which brings great computational consumption.
Therefore, we use dynamic path and relation evaluation indicators, which can constantly update during the path search and ultimately guide our path search strategy. 
Specifically, we build a path pool $\mathbf{P}_{l}$ to save the searched path.
When a path is discovered for the first time, we initialize its dynamic path indicators involving path support, path count, path confidence, and entity pair coverage based on the current head entity. 
When searching for an existing path feature, we dynamically update its indicators. 
In this way, the path pool is constantly updated during path search, in which new paths are constantly added, and existing paths are constantly updated with dynamic path indicators. The dynamic confidence and dynamic entity pair coverage are approximate as follows:
\begin{equation}
  \textsf{D-confidence}\left(P\right)\approx \frac{\sum_{i=1}^{k}  \textsf{support}\left(e_{h_{i}}, e_{t_{i}}, P\right)}{\sum_{i=1}^{k}  \textsf{count}\left(e_{h_{i}}, P\right)},
\end{equation}

\begin{equation}
  \textsf{D-coverage}\left(P\right)\approx \frac{\sum_{i=1}^{k}  \textsf{cover}\left(e_{h_{i}}, e_{t_{i}}, {P}\right)}{k},
\end{equation}
where $k$ represents the kth entity pair currently traversed. At the same time, we set a dynamically changing relation matrix $C=\{c_{ij}\}_{m\times n}\in \mathbb{R}^{m\times n}$, where $m$ is the total number of relations in the KG, $n$ is the number of paths in the current path pool, the entry $c_{ij}$ 
represents whether the current relation $r_i$ participates in the path $P_j$ as follows:
\begin{equation}
  c_{ij}=\left\{
  \begin{aligned}
    \textsf{D-confidence}\left({P_j}\right), \ & r_i~\text{in}~P_j \\
    0~~~~~~~~~~~~~~, \ & r_i~\text{not~in}~P_j\\
  \end{aligned}
  \right.
\end{equation}
The confidence vector of relation $r_i$ is denoted as:
\begin{equation}
  C_{r_i} = [c_{i1}, c_{i2}, \cdots, c_{in}].
\end{equation}
Such a relation matrix can reflect the current importance of different relations and guide path search. 
We use the following three strategies to narrow the path search space: 1) Probabilistic searching based on dynamic relation confidence, 2) Sampling the entities connected by the same relation. 3) Stopping immediately after finding any tail entity.

\begin{figure}
\includegraphics[width=\textwidth]{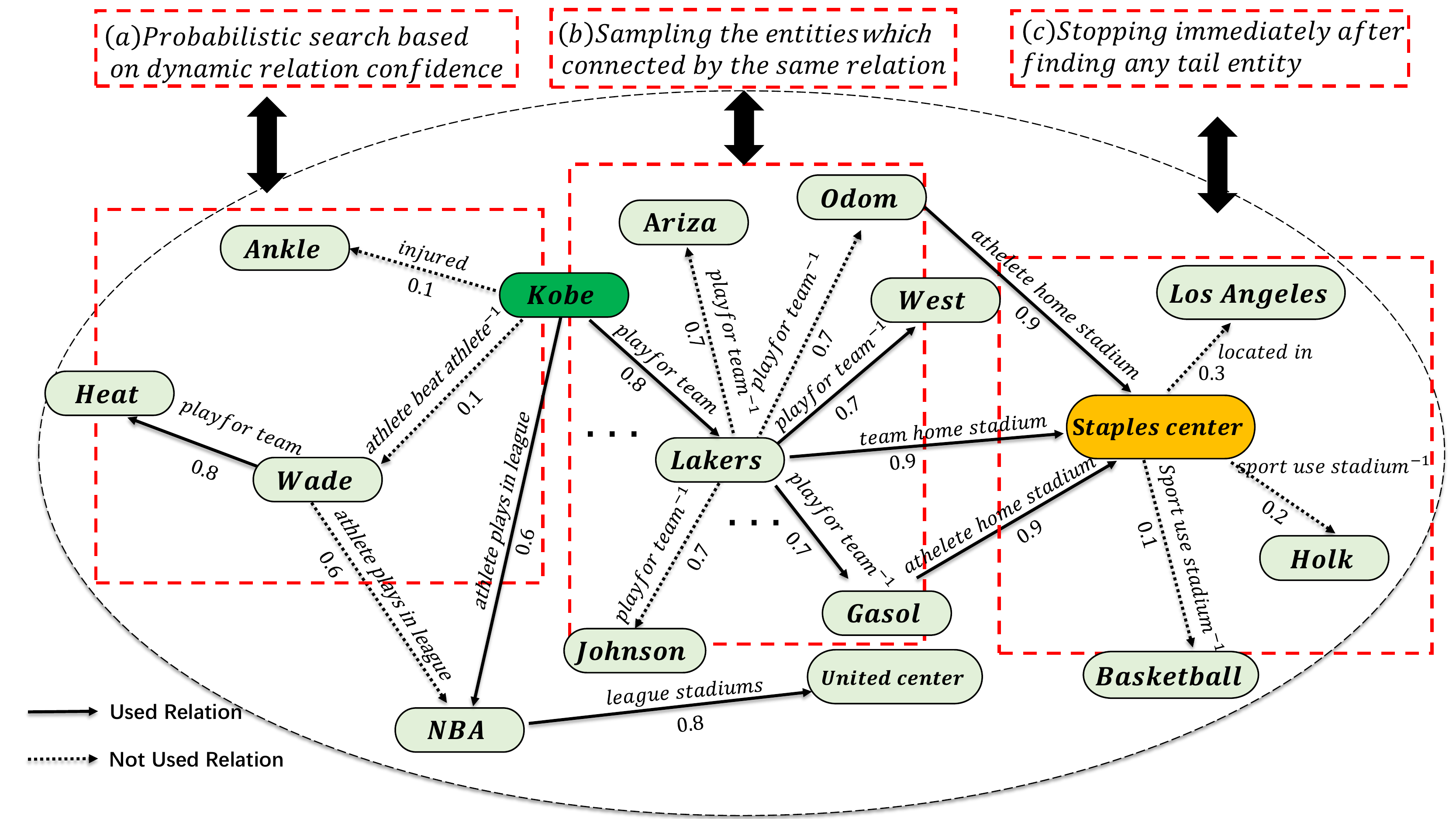}
\caption{A simple example of a path search. Reasoning task: the player’s home stadium, the starting head entity $e_h$ is Kobe Bryant, and the correct tail entity $e_t$ is Staples stadium. Three strategies are shown in the three red dotted boxes.} \label{fig:sport}
\end{figure}

Fig.~\ref{fig:sport} provides a simple example of path search in the KG, where the target task is to reason athlete's home stadium, the current head entity is the athlete: Kobe Bryant and the tail entity is the Staples stadium. 
The number marked below the relation in the figure is the probability of continuing the deep search for the relation, which will be introduced below. In general, we use such simple examples to illustrate our three path search strategies.

\subsubsection{Probabilistic search based on dynamic relation confidence}
For each relation $r_i$ that exists in the KG, the relation confidence is defined as:
\begin{equation}
A_{r_{i}}=\left(\text{Max} \left(C_{r_{i}}\right)+\text { Average }\left(C_{r_{i}}\right)\right) / 2
\end{equation}
This is the combination of the maximum confidence and the average confidence of the path that the current relation has participated in.
If the relation confidence of a relation is close to 1, it means that the average confidence and maximum confidence of the path it has participated in are both high, and this relation will be given priority when searching for the path. 
Among them, we use a probability function $prob(r_i)$:
\begin{equation}
{prob}\left(r_{i}\right)=\left(\alpha    A_{r_{i}}+\beta\right)^{\gamma},
\label{prob}
\end{equation}
which represents the probability of continuing the deep search along with the relation $r_i$. $\alpha$, $\beta$, $\gamma$ are used to control the degree of change in probability with the relation of confidence.
During the search process, such a probability function generates the probability between 0 to 1 according to the relation confidence of the current relation.
Such a probability function will ensure that we can continuously search for those important relations and terminate the search for irrelevant ones in times.
As shown in part (a) of Fig.~\ref{fig:sport}, for the reasoning task of the  stadium, the relation between the athlete's injury and the athlete's opponent hardly participates in any path feature, so they have a low search probability.
In the process of traversing entity pairs and updating the path pool, the search target gradually focus on those important relations with a high search probability, to narrow the search space and change the path feature evaluation.
Particularly, if a path contains only one relation, the search will not continue. Because such a relation may be a synonym of the target relation. We will start the probabilistic search based on relation confidence after a certain number of entity pairs are traversed. 

\subsubsection{Sampling the same relation of entities}
In order to further narrow the search space, when an entity is connected to multiple entities through the same relation, we only search for some of them by sampling. Through such sampling, the efficiency of path search can be improved without losing too much information.
As shown in Fig.~\ref{fig:sport}, Lakers have a large number of players. Therefore, we can only sample and search some of its players to approximate the overall effect. This ability strategy allows us to efficiently perform path search in the KG space with a large entity out-degree value and will not completely discard the information of these height out-degree entities.

\subsubsection{Stopping immediately after finding any tail entity}
Once we find any target entity, we will stop the path search. Although it is possible to find another target tail entity from the target entity, such a strategy can simplify the path to a certain extent.
As shown in Figure 2 (c), we will not continue to search after searching Staples stadium to avoid redundant path features. 

The probabilistic depth search algorithms we called DFS-conf are shown in Algorithm 1.
\begin{algorithm}
% \algsetup{linenosize=\scriptsize} \scriptsize

    \caption{DFS-conf}
    \KwIn{Head entity $e_h$, tail entity set $T(e_h)$, max path length $l$, out-degree threshold $\theta_o$}
    \KwOut{Path pool $\mathbf{P}_{l}$}
    Initialization $step$=0\;
    \For{relation in $e_h$'s conjunction}
    {   generate a ramdom number $a$ in [0,1]\;
        calculate search probability of  relation by formula~\ref{prob}
        
        \If {$prob(relation)<a$}
        {
        contine\;
        }
        \Else{
            entity set $\leftarrow$ entites connect with head entity by relation\;
            \If{$|$entity set$|$$>\theta_o$}
            {
            entity set $\leftarrow$ random sample $k$ entities from entity set\;
            }
            
            \If {find path}
            {
            Update $\mathbf{P}_{l}$,C\;
            contine\;
            }
            \If{step $>=$ $l$}
            {
            \Return
            }
        \For{entity in entity set}
        {
        step += 1\;
        DFS-conf(entity,$T(e_t)$)\;
        }
        }

    }
  \Return{}
  \label{alg:2}
\end{algorithm}

\subsection{Path Selection}
After path search, we select the path features in the path pool and set the entity pair coverage threshold $\theta_p$ to delete those path features with high confidence but extremely low coverage. 
Although the confidence of these path features is high, the scope of application in the knowledge is narrow. 
And we use segmentation thresholds $\theta_c$ to filter paths based on the confidence of the path.
Therefore, we usually set different thresholds for paths with a length of 1 and greater than 1.

\subsection{Train linear regression models for relation reasoning tasks}
Similar to algorithms such as PRA and DeepPath, we perform path search and get the filtered path, and finally, train a simple linear regression to perform the relation reasoning task. The input feature is a probability matrix representing the probability of the current head entity being able to walk to the correct tail entity through the current path. The higher the final output score, the more likely the entity pair has the relation.

\section {Experiments}

\subsection{Dataset}
We evaluate the proposed method on two datasets: NELL-995~\cite{xiong2017deeppath} and FB15K-237~\cite{toutanova2015representing}. 
The NELL-995 dataset is a subset of the 995th iteration of the NELL system, and the frequent but meaningless relations are deleted. The FB15K-237 dataset is a subset sampled from FB15K~\cite{bordes2013translating}. During path search, we only delete the corresponding triples and inverse triples from the KG for the currently searched head entity and tail entity set. The specific information of the datasets is shown in Table~\ref{data sets}.
\begin{table}
  \renewcommand\arraystretch{1} 
  \centering
  \caption{The basic statistics of the two knowledge graph datasets.}
  \label{data sets}
  \resizebox{0.8\textwidth}{!}{% 
  \begin{tabular}{ccccc}
  \hline

  Dataset       & entity num   & relation num &  triple num  & task num \\ 
  \hline
  NELL-995  & 75492 & 200 & 154213 & 12  \\
  FB15K-237 & 14505 & 237 & 310116 & 20  \\

  \hline
  \end{tabular}}
\end{table}

\subsection{Baseline and details}
We compare our method with the following methods: translation models, Analogy, Rescal, and DeepPath.
\begin{itemize}
 \item [$\bullet$] \textbf{Translation models.} 
   We use the following translation method as the baseline:
  
  \noindent \textbf{TransE}~\cite{bordes2013translating}. It regards the relation as the translation of the head entity to the target entity and map entities and relations to the same vector space by constraining the difference between vectors. 
  
  \noindent \textbf{TransR}~\cite{TransR}. It improves on TransE~\cite{bordes2013translating} and embeds entities and relations into different spaces. For the relation, an additional matrix $M_r$ is added to describe the space where the relation is located.
  
  \noindent \textbf{TransH}~\cite{wang2014knowledge}.  It embeds entities and relations into the same vector space, but the representation of entities in different relations is different.
  
  \noindent \textbf{TransD}~\cite{ji2015knowledge}. It simplifies the TransR~\cite{TransR} model,  uses two vectors to represent entities or relations. One of the vectors represents the entity and relation, and the other is used to construct the dynamic mapping matrix.

 \item [$\bullet$] \textbf{Rescal}~\cite{nickel2011three}. It is a semantic matching model that performs reasoning by matching the underlying semantics of entities and the relations in the vector space. Its scoring function is bilinear and uses a matrix to represent relations.
 \item [$\bullet$] \textbf{Analogy}~\cite{liu2017analogical}. It further models the analogy properties of entities and relations. At the same time, Analogy proved that DistMult~\cite{yang2014embedding}, HolE~\cite{xue2018expanding}, ComplEx~\cite{trouillon2016complex} and other models can be regarded as its special cases.

\item [$\bullet$] \textbf{DeepPath}~\cite{xiong2017deeppath}. It use reinforcement learning to search for paths and set up three different rewards functions: global accuracy rewards, path efficiency rewards, and path diversity rewards to find paths. Its reasoning performance has been proved to be better than the path ranking algorithm.

\end{itemize}

\subsection{Experimental Setting}
\subsubsection{Reasoning task}
For a head entity and target relation, we try to find the tail entity that is most likely to form a triple from the candidate tail entity.
For the relation $r_t$ to be reasoned, the set of positive samples is denoted as $D$, which is further split into training and testing sets. 

We use the training set for path search and linear model training and the reasoning performance is evaluated based on the testing set.
A positive sample and its corresponding negative samples together form a sequence. The trained linear model can calculate the score of each candidate triple in the sequence and sort the sequence in descending order. We evaluate the performance of relation reasoning according to the ranking of positive samples in the sequence using mean average precision (MAP) index: If there are $k$ pairs of correct triples in sequence, the MAP could be calculated as follows:
\begin{equation}
M A P=\frac{\sum_{i=1}^{k} r a n k\left(e_{h}, e_{t_{i}}\right)}{k}
\end{equation}
Finally, we get the average MAP of all entity pairs in the testing set.

\subsubsection{Details and parameter settings}
The code of translation models comes from \textit{Fast-TransX}\footnote[1]{\url{https://github.com/thunlp/Fast-TransX/}}. For each reasoning task, we delete the positive sample entity pairs in the testing set from the KG to perform the embedding task. The learning rate is set to 0.01 and the margin is 1, the relation and entity embedding dimensions are both 100, and the training is performed 1000 times.
For Analogy and Rescal, we get the result based on the provided code in \textit{OpenKE}\footnote[2]{\url{https://github.com/thunlp/OpenKE/}}  and use the default parameters.
The DeepPath algorithm is based on Xiong's code at \textit{DeepPath}\footnote[3]{\url{https://github.com/xwhan/DeepPath./}}.  
For our method DC-Path, the maximum path length is set to 3. In the probability function, $\alpha$, $\beta$, $\gamma$ are set to 0.99, 0.01, 0.5 respectively. Four threshold parameters are set for path selection: $\theta_{c_1}$ and $\theta_{c_2}$ represent the confidence threshold with path length equal to 1 and greater than 1 respectively. $\theta_{p_1}$ and $\theta_{p_2}$ represent the entity pairs coverage threshold with path length equal to 1 and greater than 1 respectively.
In the NELL-995 data set, we set the above four thresholds as\{0.3,0.5,0.01,0.1\} and\{0.2,0.3,0.02,0.2\} for FB15k-237. 
In FB15k-237, we are lowering the threshold to \{0.2,0.2,0.02,0.02\}.

\begin{table*}[htp]
\centering
\caption{Results of relation reasoning. The best results are marked in bold.}
\resizebox{\textwidth}{!}{%
\begin{tabular}{lcllllllcl} 
\hline\hline
\multirow{2}{*}{DateSet}    & \multirow{2}{*}{Task}          & \multicolumn{8}{c}{Method}                                                                                                    \\ 
\cline{3-10}
                            &                                & TransE & TransR          & TransD          & TransH          & Analogy & Rescal          & DeepPath        & DC-Path             \\ 
\hline
                            & agentBelongsToOrg     & 0.746  & 0.747           & 0.723           & \textbf{0.759 } & 0.708   & 0.669           & 0.576           & 0.650            \\
\multirow{12}{*}{NELL-995}  & athleteHomeStadium             & 0.711  & 0.757           & 0.656           & 0.680           & 0.751   & 0.662           & 0.848           & \textbf{0.904 }  \\
                            & athletePlaysForTeam            & 0.685  & 0.739           & 0.618           & 0.641           & 0.714   & 0.602           & 0.712           & \textbf{0.818 }  \\
                            & athletePlaysInLeague           & 0.921  & 0.814           & 0.941           & 0.919           & 0.848   & 0.911           & 0.955           & \textbf{0.975 }  \\
                            & athletePlaysSport              & 0.982  & 0.943           & 0.981           & 0.950           & 0.909   & 0.952           & 0.896           & \textbf{0.984 }  \\
                            & orgHeadquaterCity              & 0.652  & 0.711           & 0.623           & 0.616           & 0.784   & 0.566           & 0.790           & \textbf{0.803 }  \\
                            & orgHiredPerson                 & 0.707  & 0.724           & 0.710           & 0.707           & 0.726   & 0.696           & 0.745           & \textbf{0.780 }  \\
                            & bornLocation                   & 0.795  & 0.711           & 0.802           & \textbf{0.819 } & 0.807   & 0.812           & 0.742           & 0.727            \\
                            & personLeadsOrg                 & 0.766  & 0.771           & 0.765           & 0.735           & 0.796   & 0.746           & 0.780           & \textbf{0.811}   \\
                            & teamPlaysInLeague              & 0.907  & \textbf{0.933 } & 0.913           & 0.918           & 0.873   & 0.895           & 0.857           & 0.910            \\
                            & teamPlaySports                 & 0.818  & 0.881           & 0.734           & 0.799           & 0.705   & 0.733           & 0.708           & \textbf{0.886}   \\
                            & worksFor                       & 0.702  & 0.695           & 0.696           & 0.684           & 0.722   & 0.692           & 0.700           & \textbf{0.743}   \\ 
\cline{2-10}
                            & Average                        & 0.783  & 0.785           & 0.764           & 0.769           & 0.779   & 0.745           & 0.776           & \textbf{0.833}   \\ 
\hline
\multirow{21}{*}{FB15k-237} & teamSport                      & \textbf{0.968}  & 0.967           & 0.939           & 0.931           & 0.891   & 0.972 & 0.868           & 0.963            \\
                            & birthPlace                     & 0.411  & 0.390           & 0.383           & 0.386           & 0.398   & 0.310           & \textbf{0.510}  & 0.441            \\
                            & personNationality              & 0.662  & 0.719           & 0.493           & 0.664           & 0.681   & 0.773           & 0.840           & \textbf{0.842}   \\
                            & fimDirector                    & 0.458  & 0.470           & 0.448           & 0.439           & 0.452   & 0.393           & 0.358           & \textbf{0.490}   \\
                            & filmWriteenBy                  & 0.623  & 0.625           & \textbf{0.642 } & 0.628           & 0.571   & 0.570           & 0.558           & 0.493            \\
                            & filmLanguage                   & 0.546  & 0.553           & 0.424           & 0.523           & 0.494   & 0.642           & 0.691           & \textbf{0.705}   \\
                            & tvLanguage                     & 0.955  & 0.960           & 0.956           & 0.942           & 0.918   & 0.954           & 0.964           & \textbf{0.967}   \\
                            & capitalOf                      & 0.520  & 0.539           & 0.541           & 0.556           & 0.527   & 0.501           & 0.743           & \textbf{0.837}   \\
                            & orgFounded            & 0.383  & 0.388           & 0.383           & \textbf{0.451 } & 0.444   & 0.375           & 0.302           & 0.279            \\
                            & musicianOrigin                 & 0.426  & 0.434           & 0.423           & 0.416           & 0.484   & 0.385           & \textbf{0.506}  & 0.446            \\
                            & serviceLocation                           & 0.483  & 0.514           & 0.530           & 0.541           & 0.523   & 0.470           & \textbf{0.556}  & 0.492            \\
                            & filmCountry                    & 0.610  & 0.565           & 0.450           & 0.584           & 0.630   & 0.644           & 0.693           & \textbf{0.708}   \\
                            & filmMusic                      & 0.507  & 0.500           & 0.526           & 0.499           & 0.384   & \textbf{0.538 } & 0.251           & 0.465            \\
                            & orgHeadquarters                & 0.580  & 0.584           & 0.606           & 0.591           & 0.422   & 0.503           & \textbf{0.616 } & 0.415            \\
                            & orgMember                      & 0.437  & 0.437           & 0.443           & 0.441           & 0.444   & 0.389           & 0.261           & \textbf{0.457}   \\
                            & professionSpecializationOf   & 0.484  & 0.478           & 0.448           & 0.464           & 0.466   & \textbf{0.607 } & 0.485           & 0.425            \\
                            & languagesSpoken                & 0.404  & 0.415           & 0.461           & 0.417           & 0.405   & 0.327           & 0.402           & \textbf{0.421}   \\
                            & timeEventLocations             & 0.355  & 0.386           & \textbf{0.395 } & 0.329           & 0.307   & 0.370           & 0.431           & 0.350            \\
                            & tvProgramGenre               & 0.401  & 0.369           & 0.386           & 0.395           & 0.412   & 0.340           & \textbf{0.511 } & 0.438            \\
                            & tvProgramCountryOfOrigin & 0.886  & 0.837           & 0.913           & 0.859           & 0.904   & 0.853           & 0.878           & \textbf{0.915}   \\ 
\cline{2-10}
                            & Average                        & 0.555  & 0.556           & 0.539           & 0.553           & 0.538   & 0.546           & 0.571           & \textbf{0.576}   \\
\hline\hline
\label{MAP}
\end{tabular}}
\end{table*}

\subsection{Results and analysis}
\subsubsection{Relation Reasoning Accuracy And Paths Used}
In this part, we report the reasoning results of each method and the analysis of the used path number.Table~\ref{MAP} reports the results of performance comparison between our method and baselines, from which we can see that DC-Path significantly outperforms most baselines in most cases.
The reasoning results in the NELL-995 data set show that in most tasks, DC-Path can get higher reasoning accuracy than these baselines. Although the reasoning results on FB15k-237 show that not every task performs better than these baselines, the average accuracy of the method is still leading. Therefore, we can conclude that DC-Path can get better reasoning accuracy than these mainstream reasoning methods.

Table~\ref{path_num} shows the number of paths used in the final reasoning of DeepPath and DC-Path. 
DC-Path greatly reduces the number of path features used and achieve better performance. 
%The number of paths ultimately used by DC-Path is not much different from that of DeepPath. Note that DC-Path only uses the path features within 3 and achieve such a good reasoning performance. This shows that our path search and selection strategy based on path confidence and entity pairs coverage is effective. 
Meanwhile, we also found that there are more path features in FB15k-237. On average, there are nearly 33 path features for each task. And the number of path features of different reasoning tasks is very unbalanced.
There are nearly a hundred paths for task: $personNational$ but only 1 path for task: $orgFounded$. This is why in sparse reasoning tasks, path-based reasoning methods are usually inferior to representation learning methods.
%At the same time, there are only about 1/5 of the number of entities in the FB15K-237 compared to NELL-995, but nearly twice the number of triples. So the number of path features between an entity pair is rise significant for some target tasks.

\subsubsection{Time Consumption Of Path Search}
We ran our code on a computer with 16GB of RAM and an i7 8th generation processor. For the NELL-995 dataset, the average time consumption on path search for each target relation is about 21s, while for FB15k-237, the average time is about 1100s. 
%The main reason for this gap is: When the target task appears in the KG more frequently, there are more positive entity pairs that need to be traversed such as task: $personPlaceOfBirth$. 
We also discover that DeepPath usually spends more search time. Since the reinforcement learning strategy of the DeepPath algorithm can discover longer path features, its path search cost is also greater, usually more than 10000 seconds. The reason may be that it needs to train the neural network many times. Therefore, our method can greatly reduce the time consumption in path search and use path features with a limited length to achieve good reasoning results.

\begin{table}[htp]
  \centering\caption{Comparison of the number of paths}
  \label{path_num}
  \resizebox{\textwidth}{!}{% 
  \begin{tabular}{cccccccc} 
  \hline\hline
  \multirow{2}{*}{Dataset}    & \multirow{2}{*}{Task}    & \multicolumn{2}{c}{Method}  & \multirow{2}{*}{Dataset}    & \multirow{2}{*}{Task}    & \multicolumn{2}{c}{Method} \\
  \cline{3-4}\cline{7-8}
                              &                            & DeepPath &DC-Path  &                            &                            & DeepPath &DC-Path    \\
  \hline 
  \multirow{20}{*}{FB15k-237} & teamSport                  & 17       & 11      & \multirow{20}{*}{NELL-995} & agentBelongstoorg          & 15       & 19                        \\
                              & birthPlace                 & 4        & 8       &                            & athleteHomeStadium         & 9        & 3               \\
                              & personNationality          & 86       & 52      &                            & athletePlaysForTeam        & 23       & 2             \\
                              & ﬁlmDirector                & 2        & 2       &                            & athletePlaysInLeague       & 31       & 34    \\
                              & filmWriteenBy              & 6        & 2       &                            & athletePlaysSport          & 15       & 21       \\
                              & ﬁlmLanguage                & 53       & 94      &                            & orgHeadquaterCity          & 5        & 19           \\
                              & tvLanguage                 & 44       & 101     &                            & orgHiredPerson             & 9        & 12            \\
                              & capitalOf                  & 3        & 6       &                            & bornLocation               & 5        & 1            \\
                              & orgnFounded                & 2        & 1       &                            & personLeadsOrg             & 15       & 9             \\
                              & musicianOrigin             & 17       & 2       &                            & teamPlaysInLeague          & 8        & 27             \\
                              & serviceLocation~           & 52       & 16      &                            & teamPlaySports             & 10       & 17             \\
                              & filmCountry                & 54       & 113     &                            & worksFor                   & 15       & 11         \\
                              & filmMusic                  & 2        & 35      &                            & /                          &          &            \\
                              & orgHeadquarters            & 8        & 2       &                            & /                          &          &      \\
                              & orgMember                  & 8        & 60      &                            & /                          &          &      \\
                              & professionSpecializationOf & 5        & 1       &                            & /                          &          &     \\
                              & languagesSpoken            & 9        & 1       &                            & /                          &          &     \\
                              & timeEventLocations         & 10       & 4       &                            & /                          &          &     \\
                              & tvProgramGenre             & 51       & 3       &                            & /                          &          &     \\
                              & tvProgramCountryOfOrigin   & 45       & 137     &                            & /                          &          &     \\ 
  \hline
                              & Average                    & 13.3     & 14.5    &                            & Average                   &23.9       & 32.6         \\ 
  \hline\hline
  % \multirow{FB15k-237}        
  \end{tabular}}
  \end{table}

\subsubsection{The Impact Of Different Confidence Thresholds On The Results}
\begin{figure}
\includegraphics[width=1\textwidth,height=0.5\textwidth]{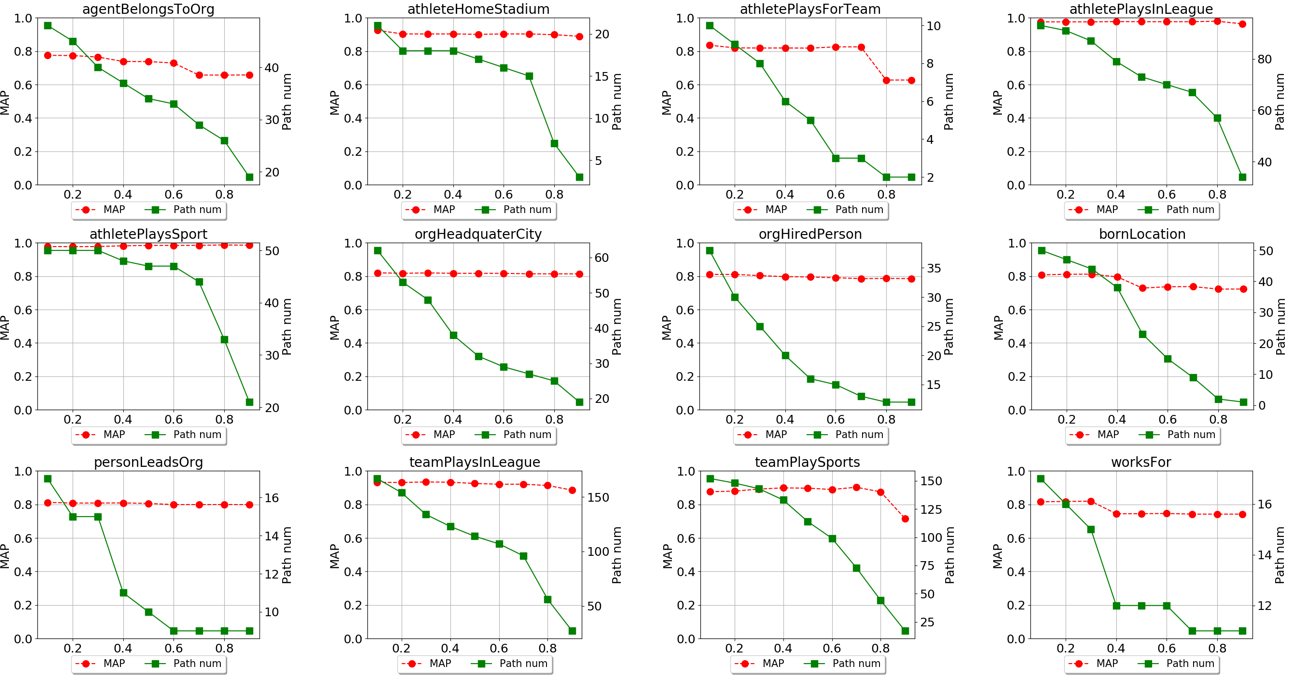}
\caption{MAP and path number in different path confidence threshold in NELL-995.} \label{fig:NELL-all}
\end{figure}

\begin{figure}
\includegraphics[width=1\textwidth,height=1\textwidth]{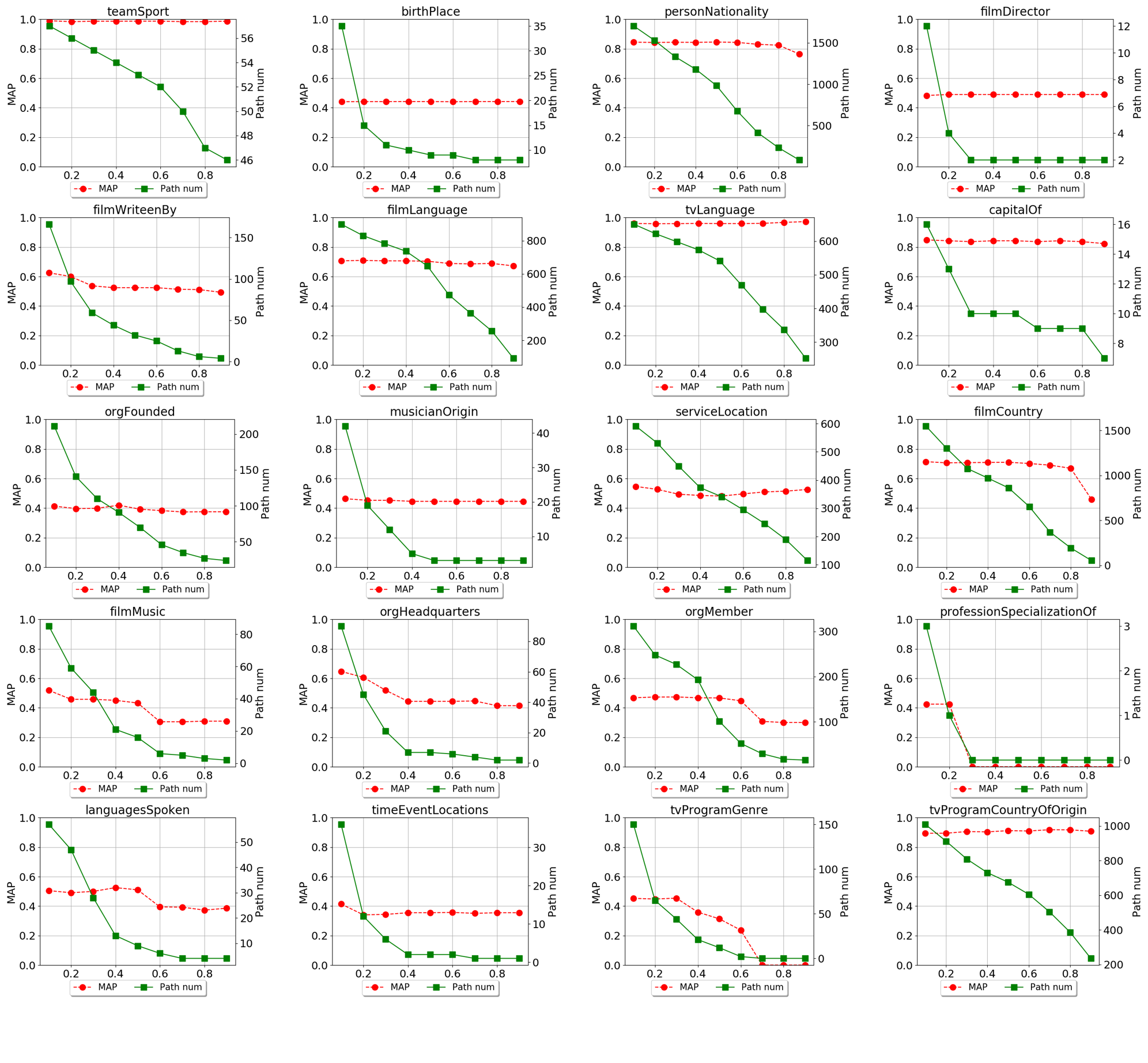}
\caption{MAP and path number in different path confidence threshold in FB15k-237.} \label{fig:FB15K-all}
\end{figure}

In this part, we explore the influence of different path confidence thresholds on the reasoning results. We keep the paths whose path length is one and the path confidence is greater than 0.3, and the entity pair coverage is greater than 0.01. On this basis, we fix the entity pair coverage threshold to 0.01 and test the accuracy of relation reasoning with different confidence thresholds. Fig.~\ref{fig:NELL-all} and Fig.~\ref{fig:FB15K-all} shows the number of path features retained by different path confidence thresholds and the reasoning accuracy (MAP) when using these path features for relation reasoning. 
% \uwave{We set the coverage threshold of all entities to 0.01 and retain the path feature of length one then take different path confidence thresholds to observe the number of retained path features and their inference accuracy.} 
% Generally speaking, the more path features are retained, the higher the inference accuracy and the more time-consuming inference. 
We can find that the MAP of many target relations does not decrease significantly as the threshold increases, which means that those paths whose path confidence is greater than 0.5 or even higher play a key role in relation reasoning.The performance of some reasoning tasks is poor because of that there are too few effective reasoning paths that meet the confidence threshold.

\subsubsection{Display Some Paths And Relations}
In this part, we show some high confidence paths in several tasks and the most relevant top-2 relations among them.
Table~\ref{path_num} shows the important paths and their confidence in some tasks.
These paths are of high quality in terms of semantic logic analysis and path confidence, which shows the effectiveness of our path search method. Through the first two most relevant relationships in the target task. It can be found that they are usually semantically closely related to the target reasoning relation, which also shows that our strategy of narrowing the path search space through dynamic relation confidence is effective. We can find that some short paths have high confidence. This proves that short paths play a greater role in reasoning.

\section{Conclusion}
This paper designs and implements a path feature search method to perform KG reasoning tasks named DC-Path. It is based on dynamic path confidence and uses it to perform relation reasoning tasks. Different from the previous reasoning methods based on graph structure , this method uses dynamic relation confidence and other indicators to achieve path search and evaluation during the searching process. Relation reasoning tasks on NELL-995 and FB15K-237 show that this method can effectively search path features for reasoning, and obtain good reasoning results, which provides new ideas for the KG reasoning method based on path features.

\subsubsection{Acknowledgment}
This work was supported by the National Natural Science Foundation of China (Grant No.62103374), Basic Public Welfare Research Project of Zhejiang Province(Grant No.LGF20F020016), Open Project of the Key Laboratory of Public Security Informatization Application Based on Big Data Architecture(Grant No.2020DSJSYS003) and Key R\&D Projects in Zhejiang Province (Grant No. 2021C01117).

\bibliographystyle{splncs04_}
\bibliography{mybibliography}

\end{document}